\documentclass[lettersize,journal]{IEEEtran}
\usepackage{amsmath,amsfonts}
\usepackage{algorithmic}
\usepackage{algorithm}
\usepackage{array}
\usepackage[caption=false,font=normalsize,labelfont=sf,textfont=sf]{subfig}
\usepackage{textcomp}
\usepackage{stfloats}
\usepackage{url}
\usepackage{verbatim}
\usepackage{graphicx}
\usepackage{cite}
\usepackage{amssymb}
\usepackage{booktabs}
\usepackage{multirow}
\usepackage[table]{xcolor}
\usepackage{graphicx}
\usepackage{subcaption}

\begin{document}

\title{TIGA: Trajectory-Injected Generative Attack against Black-box AIGC Detectors}

\author{{Xia Du, Zhuosen Bao, Zheng Lin, Jizhe Zhou, Chi-man Pun,~\IEEEmembership{Senior Member,~IEEE}, Jun Luo,~\IEEEmembership{Fellow,~IEEE}, and Symeon Chatzinotas,~\IEEEmembership{Fellow,~IEEE}
\thanks{Xia Du and Zhuosen Bao are with the School of Computer and Information Engineering, Xiamen University of Technology, Xiamen, 361000, China (email: baozhuosen@stu.xmut.edu.cn; duxia@xmut.edu.cn; szzhu@xmut.edu.cn).}
\thanks{Zheng Lin and Symeon Chatzinotas are with the Interdisciplinary Centre for Security, Reliability and Trust (SnT), University of Luxembourg, L-1855 Luxembourg, Luxembourg (e-mail: zhenglin@ieee.org; symeon.chatzinotas@uni.lu).}
\thanks{Jizhe Zhou is with the School of Computer Science, Engineering Research Center of Machine Learning and Industry Intelligence, Sichuan University, Chengdu, China, 610020, China (email: jzzhou@scu.edu.cn).}
% \thanks{Jiawei Lian is with the PCA Laboratory, Key Laboratory of Intelligent Perception and Systems for High-Dimensional Information of Ministry of Education, School of Computer Science and Engineering, Nanjing University of Science and Technology, Nanjing 210094, China (e-mail: lianjw@njust.edu.cn).}
\thanks{Chi-man Pun is with the Department of Computer and Information Science, Faculty of Science and Technology, University of Macau, Macau, 999078, China (email: cmpun@umac.mo).}}
% \thanks{Wei Ni is with the School of Engineering, Edith Cowan University, Perth, WA 6027, Australia (email: wei.ni@ieee.org).}
\thanks{Jun Luo is with the College of Computing
and Data Science, Nanyang Technological University, Singapore (e-mail: junluo@ntu.edu.sg).}

\thanks{Corresponding authors: Zheng Lin (zhenglin@ieee.org).}
}


\maketitle

\begin{abstract}
Recent diffusion models have achieved remarkable realism in facial image synthesis, posing growing challenges to artificial intelligence-generated content (AIGC) forensic detectors. Existing evasion methods typically perturb pre-generated images or require detector-aware training, which may introduce visible or statistical artifacts and limit applicability when the diffusion model must remain frozen and the target detector is accessible only through black-box queries. We propose Trajectory-Injected Generative Attack (TIGA), a source-image-free and training-free framework that generates detector-evasive images within a single diffusion sampling trajectory. TIGA steers the latent Denoising Diffusion Implicit Model (DDIM) trajectory so that adversarial properties emerge during generation rather than being added afterward. TIGA first aggregates gradients from multiple white-box surrogate detectors to form a transferable, sign-aware prior, and then performs anisotropic directional search with symmetric finite-difference queries to estimate the black-box target response. The estimated directions are stabilized by decayed momentum and injected according to the DDIM noise schedule, with frequency-domain reshaping to suppress high-frequency artifacts. Experiments on surrogate and unseen specialized forensic detectors show that TIGA achieves strong black-box attack performance, transferability, and high robustness under common post-processing operations without source images or diffusion-model retraining, while preserving high perceptual quality.
\end{abstract}


\section{Introduction}
\IEEEPARstart{R}{ecent} diffusion models~\cite{ho2020denoising, rombach2022high, huang2023collaborative} have achieved remarkable progress in image synthesis, enabling the generation of highly realistic artificial intelligence (AI)-generated images that are increasingly difficult to distinguish from real images by human observers. Benefiting from their powerful generative capability, diffusion-based Artificial Intelligence Generated Content (AIGC) technologies have been widely adopted in digital content creation, virtual avatars, image editing, and multimedia generation. However, the rapid proliferation of realistic AIGC has raised serious concerns regarding misinformation, malicious forgery, identity manipulation, and media credibility. To mifigate these risks, a wide range of AIGC forensic detectors have been proposed, including Convolutional Neural Network (CNN)-based detectors that learn manipulation-sensitive spatial or residual cues~\cite{bayar2016deep,wang2020cnn}, vision-transformer and foundation-model-based detectors that exploit pretrained visual representations~\cite{ojha2023towards,li2024unionformer}, frequency-domain methods that capture spectral artifacts introduced by image synthesis pipelines~\cite{qian2020thinking,tan2024rethinking}, and diffusion-reconstruction-based approaches that distinguish real and generated images through reconstruction discrepancies~\cite{wang2023dire,luo2024lare}.

% including CNN-based detectors~\cite{bayar2016deep, wang2020cnn, liu2020global, chen2021image, tan2023learning}, vision transformer detectors~\cite{ojha2023towards, yan2025sanity, yan2024orthogonal, li2024unionformer}, frequency-domain forensic methods~\cite{qian2020thinking, cai2021freqnet, yan2026dual, tan2024rethinking}, and diffusion-reconstruction-based approaches~\cite{wang2023dire, ricker2024aeroblade, luo2024lare, chen2024drct}.

% hussain2020adversarial,
Recent studies have shown that AIGC detectors remain vulnerable to adversarial evasion. Existing adversarial evasion methods can be generally divided into two categories. The first category performs post-hoc perturbation on already generated images, where adversarial signals are optimized in the pixel domain, frequency domain, or through realistic post-processing operations after image synthesis~\cite{carlini2020evading, diao2026vulnerabilities,xie2025take}. Although such perturbations are typically bounded by an $\ell_p$ budget or a trade-off parameter, these controls only limit their magnitude without keeping them on the diffusion manifold: tightening them trades attack strength for quality along a fixed frontier, and the residual may still cause visible artifacts, abnormal frequency patterns, or degraded quality within the budget, while remaining sensitive to image transformations and post-processing.

\IEEEpubidadjcol
The second category moves beyond direct post-hoc perturbation by leveraging diffusion models for detector-evasive image generation or refinement~\cite{zhou2024stealthdiffusion,wang2026adversarial}. By optimizing detector-evasive objectives within diffusion-based pipelines, these methods can alleviate the artifact problem caused by post-hoc perturbation and improve the visual fidelity of adversarial images. However, existing methods typically require detector-aware retraining, partial finetuning of diffusion components, or post-generation optimization initialized from pre-existing source images, making them difficult to apply in source-free black-box scenarios where no original image is available and the diffusion model must remain frozen.

\begin{figure*}
\centering
\includegraphics[width=0.92\linewidth]{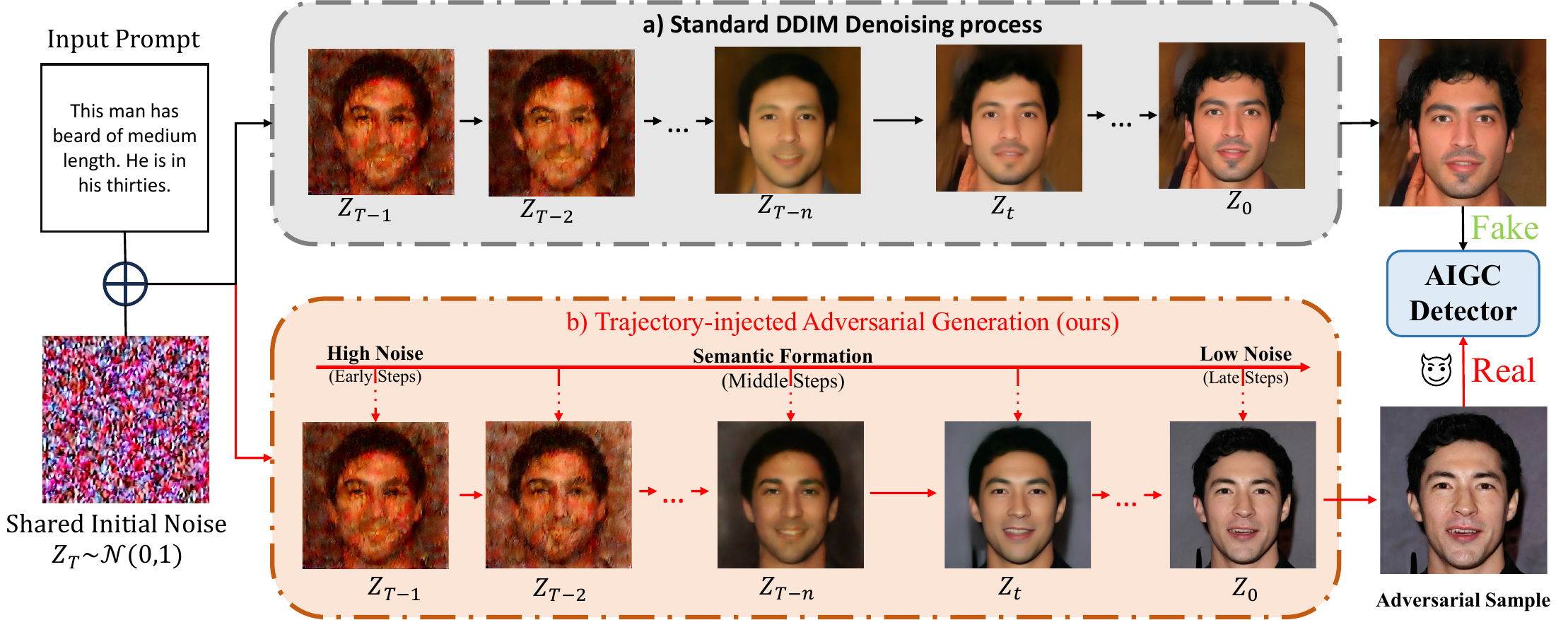}
\caption{Conceptual principle of TIGA. Standard DDIM sampling generates visually realistic AIGC images that can still be identified as fake by forensic detectors. In contrast, TIGA introduces trajectory-level adversarial guidance during diffusion sampling. Surrogate detectors first provide a transferable prior direction, while black-box symmetric probing refines the trajectory direction according to the target detector response. The resulting schedule-aware and frequency-shaped injection shifts detector decisions from fake to real while preserving visual quality and perceptual naturalness. EMA denotes exponential moving average.}
\label{fig:motivation}
\end{figure*}

In practice, a more realistic threat model arises when attackers can query the output confidence of a target AIGC detector but cannot access its internal architecture, parameters, or gradients. Meanwhile, several accessible surrogate detectors may provide transferable forensic priors. Under this setting, a key question remains underexplored: can detector-evasive images be generated directly within a single diffusion sampling trajectory, starting only from random noise and conditional inputs, without relying on pre-existing source images or retraining the diffusion model? This source-image-free black-box setting is challenging for two reasons. On the one hand, direct zeroth-order search in the high-dimensional latent space incurs prohibitive query costs and often yields unstable direction estimates. On the other hand, naively injecting adversarial guidance into the denoising trajectory may distort the generated content or introduce unnatural high-frequency artifacts, particularly when the guidance is inconsistent with the diffusion noise schedule.

To address these challenges, we propose Trajectory-Injected Generative Attack (TIGA), a training-free trajectory-injected adversarial generation framework against black-box AIGC detectors via surrogate-guided directional search. Instead of perturbing generated images after synthesis, TIGA directly steers the Denoising Diffusion Implicit Model (DDIM) denoising trajectory in the latent space, allowing detector-evasive properties to emerge during the generation process. As illustrated in Fig.~\ref{fig:motivation}, TIGA shifts the detector decision by modifying the intermediate sampling trajectory rather than attaching external perturbations to the final image, thereby keeping the result a naturally generated sample rather than an artifact-laden image. The diffusion model remains completely frozen throughout sampling, and the target detector is only queried in a black-box manner. 

The key idea of TIGA is to combine transferable white-box surrogate guidance with prior-guided black-box directional search, thereby avoiding detector-aware retraining and undirected isotropic zero-order exploration. Specifically, TIGA consists of three tightly coupled modules. First, a Surrogate-Guided Prior (SGP) module constructs a transferable adversarial direction by aggregating latent-space gradients from multiple white-box surrogate detectors. This prior direction provides both transferable detector-evasive knowledge and a sign-aware search orientation, enabling the subsequent black-box search to start from a more informative region rather than from random isotropic exploration. Second, a Surrogate-Guided Directional Search (SGDS) module performs black-box trajectory guidance under an anisotropic direction distribution centered around the surrogate prior. By combining the surrogate-guided direction with orthogonal exploration components, SGDS estimates the response of the black-box detector through symmetric finite-difference queries and aggregates the response-weighted probing directions into a stable trajectory guidance vector. This design generalizes conventional isotropic zero-order search as a boundary case while significantly reducing estimation variance under a fixed query budget. Third, a Schedule-Aware Trajectory Injection (SATI) module injects the accumulated momentum into the DDIM trajectory according to the diffusion noise schedule. The injected update is further reshaped in the frequency domain to suppress high-frequency artifacts and remove channel-wise bias, improving visual fidelity while weakening detector-sensitive forensic traces.

Different from post-hoc perturbation methods, TIGA does not attach an external adversarial pattern to the final image. Instead, it modifies the intermediate denoising trajectory so that the final sample remains naturally generated by the diffusion model, gaining evasiveness without the pixel-space residual that forces post-hoc methods to trade attack strength for quality. Unlike retraining-based adversarial generation methods, TIGA does not require optimizing or finetuning any component of the diffusion model. Moreover, compared with purely random black-box search, the proposed surrogate-guided anisotropic search yields more stable and reliable direction estimates by constraining the exploration space around transferable forensic directions.

The main contributions of this paper are summarized as follows:
\begin{itemize}
\item{We propose TIGA, a training-free trajectory-injected adversarial generation framework against black-box AIGC detectors. Without retraining diffusion models or accessing gradients of the target detector, TIGA directly manipulates DDIM denoising trajectories during sampling, allowing detector-evasive properties to emerge as part of the generation process.}

\item{We integrate prior-guided black-box search into the diffusion sampling trajectory: gradients from multiple white-box surrogate detectors are aggregated into a transferable, sign-aware prior  that only biases the search direction, while the black-box queries determine the actual update; the symmetric finite-difference query estimates are accumulated across steps into a stable trajectory guidance signal, coupling the per-step search with the generation process rather than attacking a fixed image.}

\item{We develop a schedule-aware trajectory injection strategy with frequency-domain artifact suppression. The proposed injection adaptively scales adversarial guidance according to the DDIM noise schedule and reshapes the injected update in the frequency domain, thereby balancing attack effectiveness, perceptual naturalness, and visual fidelity.}
\end{itemize}

\section{Related Work}

\subsection{AIGC Image Detection}
% including CNN-based detectors~\cite{bayar2016deep, wang2020cnn, liu2020global, chen2021image, tan2023learning}, vision transformer detectors~\cite{ojha2023towards, yan2025sanity, yan2024orthogonal, li2024unionformer}, frequency-domain forensic methods~\cite{qian2020thinking, cai2021freqnet, yan2026dual, tan2024rethinking}, and diffusion-reconstruction-based approaches~\cite{wang2023dire, ricker2024aeroblade, luo2024lare, chen2024drct}.

With the rapid development of generative models, especially generative adversarial networks (GANs)~\cite{karras2017progressive, karras2019style} and diffusion models~\cite{rombach2022high, huang2023collaborative, wang2025camd}, detecting AI-generated images has become an important problem in multimedia forensics. Early studies mainly focused on detecting images synthesized by GAN-based generators, where CNNs were trained to capture spatial artifacts and statistical inconsistencies left by image synthesis models ~\cite{bayar2016deep, wang2020cnn, liu2020global, chen2021image}. These methods demonstrated that generated images often contain model-specific or generator-agnostic forensic traces, making it possible to distinguish real and fake images even when the visual appearance is highly realistic.

Beyond spatial-domain artifacts, many works have explored frequency-domain cues for AIGC image detection. Since generative models may introduce abnormal spectral distributions, periodic patterns, or high-frequency inconsistencies during image synthesis, frequency-aware detectors attempt to identify fake images by analyzing Fourier spectra, local frequency statistics, or texture-level forensic traces ~\cite{qian2020thinking, yan2026dual, tan2024rethinking, tan2024frequencyaware}. These methods are particularly relevant to diffusion-generated images, as subtle frequency fingerprints may remain even when pixel-level visual artifacts are difficult to observe. However, frequency-based detectors may be sensitive to image compression, resizing, and adversarial frequency manipulation.

Recently, transformer-based and foundation-model-based detectors
have been introduced to improve the generalization ability of AIGC
detection. Compared with conventional CNN-based detectors, vision
transformers and Contrastive Language--Image Pre-training
(CLIP)-based representations can capture more global
semantic and structural cues, which may help detect images generated
by unseen models or under cross-domain settings
~\cite{ojha2023towards,yan2025sanity,li2024unionformer,du2026can}.
In addition, recent studies and benchmarks have systematically examined
fake-image detection and localization across different generator
families, image domains, post-processing conditions, and increasingly
realistic manipulation scenarios
~\cite{ma2024imdl,du2026forensichub,li2026artificial,
zhu2023genimage,zhu2026revisiting}.
These studies show that although existing forensic models achieve
promising performance on in-distribution test data, their robustness
and generalization can degrade significantly under unseen generators,
cross-domain data, image transformations, or complex manipulation
scenarios.

Another line of research detects diffusion-generated images based on reconstruction behavior. For example, diffusion-reconstruction-based methods observe that images generated by diffusion models can often be reconstructed more faithfully by a pretrained diffusion model than real images, and therefore use reconstruction errors as forensic representations~\cite{wang2023dire,ricker2024aeroblade,luo2024lare,chen2024drct}. Such methods provide a new perspective for identifying diffusion-generated content and have shown strong detection performance in several settings. Since these detectors rely on specific reconstruction discrepancies or forensic traces, they may still be vulnerable when the generation trajectory is intentionally steered to suppress detector-sensitive cues.

These existing AIGC detectors exploit diverse forensic evidence, including spatial artifacts, frequency statistics, semantic representations, and diffusion reconstruction errors. However, the increasing realism of diffusion-generated images and the emergence of adversarial evasion methods raise new concerns about the reliability of these detectors. This motivates the study of adversarial generation methods that can directly challenge AIGC detectors under realistic black-box settings.

\subsection{Adversarial Attacks against AIGC Image Detectors}

Adversarial attacks were initially studied in conventional image classification, where carefully optimized perturbations were introduced to cause erroneous model predictions. Representative white-box methods include Fast Gradient Sign Method (FGSM), Projected Gradient Descent (PGD), and the C\&W attack~\cite{goodfellow2014explaining,madry2018towards,carlini2017towards}. When the target model is inaccessible, transfer-based attacks craft adversarial examples using surrogate models and exploit cross-model transferability~\cite{papernot2017practical,dong2018boosting,du2025defensive}, whereas query-based attacks estimate adversarial directions directly from model outputs through zeroth-order optimization, evolutionary strategies, or random search~\cite{guo2019simple,andriushchenko2020square,vo2024brusleattack,zhu2024dp}. Although query-based attacks avoid access to target gradients, estimating reliable directions in high-dimensional image spaces often requires a substantial number of model queries. To improve query efficiency, a line of prior-guided black-box attacks combines the two paradigms using surrogate gradients as a search prior: Guided ES augments random search with surrogate-gradient directions to shape the sampling distribution~\cite{maheswaranathan2019guided}, and P-RGF further fuses a transfer-based prior with symmetric finite-difference queries to the target model, substantially reducing the query cost of gradient estimation~\cite{cheng2019improving}. Our surrogate-guided directional search builds on this prior-guided paradigm and extends it from perturbing a fixed image to steering a diffusion sampling trajectory.

Recent studies have extended adversarial evasion to DeepFake and AIGC image detectors. Early work demonstrated that synthetic-image detectors can be bypassed under both white-box and black-box settings using carefully optimized additive perturbations~\cite{carlini2020evading,hussain2021adversarial}. Subsequent methods increasingly exploit forensic properties specific to generated images. StatAttack minimizes the statistical discrepancy between real and fake images by adversarially optimizing natural degradations such as exposure, blur, and noise~\cite{hou2023evading}. FPBA introduces frequency-domain perturbations and a post-train Bayesian surrogate strategy to improve transferability across heterogeneous AIGC detectors~\cite{diao2026vulnerabilities}. R$^2$BA performs score-based black-box optimization over realistic post-processing operations, including Gaussian blur, JPEG compression, Gaussian noise, and illumination effects~\cite{xie2025take}. 

Most of these methods follow a post-hoc paradigm: they require a pre-generated fake image and optimize pixel-domain, frequency-domain, or post-processing transformations to change the detector prediction. Since the adversarial modification is performed after image synthesis, the resulting signal is not inherently produced by the original generative trajectory. Conventional perturbation-based attacks may therefore exhibit limited visual invisibility, cross-detector transferability, or robustness to practical image processing, motivating attacks that explicitly optimize natural degradations and low-level forensic statistics~\cite{hou2023evading,diao2026vulnerabilities,xie2025take}.

StealthDiffusion represents an intermediate direction between post-hoc perturbation and generation-time adversarial synthesis. Given an initially generated image, it performs latent adversarial optimization and employs a control-oriented variational autoencoder (VAE) to reduce visual and spectral discrepancies, thereby producing higher-quality detector-evasive images~\cite{zhou2024stealthdiffusion}. Nevertheless, it remains a diffusion-assisted post-generation refinement pipeline that depends on a pre-existing generated image. More recently, the Adversarial Diffusion Model (ADM) directly generates detector-evasive images from scratch by introducing an adversarial denoising U-Net to search for detector-sensitive latent representations~\cite{wang2026adversarial}. Although ADM moves adversariality into the generation process, it requires additional detector-aware optimization of diffusion components, coupling the resulting generator to the detectors used during model construction.

Different from these approaches, TIGA considers a source-image-free, training-free, and score-based black-box adversarial generation setting. It does not require an externally provided source image, post-generation refinement, or parameter updates to the diffusion model. TIGA uses accessible surrogate detectors only to construct a transferable directional prior and employs confidence queries to the black-box target detector to refine the adversarial direction online. The resulting guidance is injected directly into the DDIM denoising trajectory, allowing detector-evasive properties to emerge during inference while keeping the diffusion model frozen.

\begin{figure*}
\centering
\includegraphics[width=0.92\linewidth]{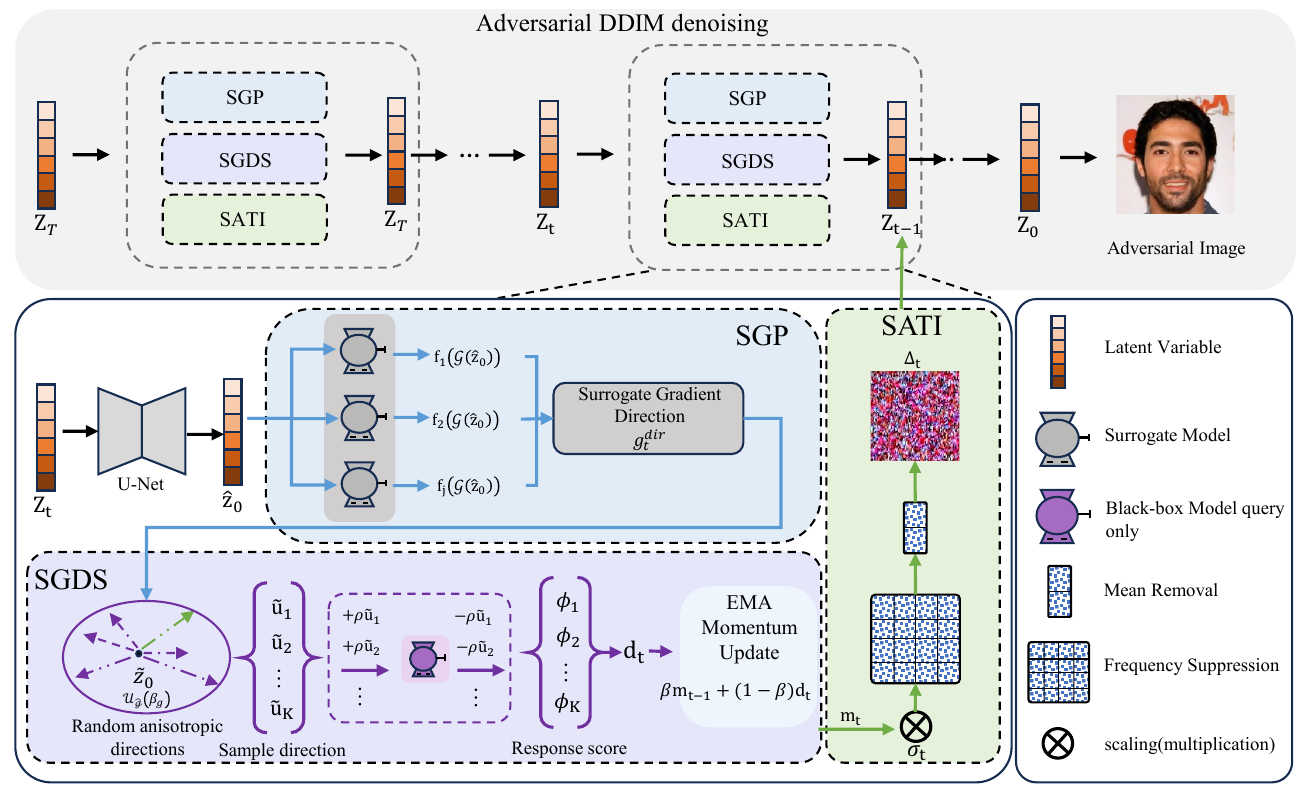}
\caption{Overall framework of TIGA. Given the shared initial noise $z_T$, standard DDIM sampling first predicts the clean latent $\hat{z}_0$ at each denoising step. The SGP module aggregates gradients from multiple surrogate detectors to construct a transferable prior direction. Guided by this prior, the SGDS module samples anisotropic probing directions, queries the black-box detector with symmetric finite differences, and accumulates the response-weighted directions into momentum. Finally, the SATI module performs schedule-aware trajectory injection with frequency-domain artifact suppression, producing detector-evasive images without retraining the diffusion model.}
\label{fig:framework}
\end{figure*}

\section{Method}
\label{sec:method}

\subsection{Overview and Threat Model}

\textbf{Overview.} As illustrated in Fig.~\ref{fig:framework}, TIGA performs adversarial generation entirely within the DDIM sampling process, without modifying or finetuning the diffusion model. Given a condition $c=(\text{mask},\text{text})$ and an initial noise $z_T$, the frozen diffusion model produces a denoising trajectory $z_T\!\rightarrow\!z_{T-1}\!\rightarrow\!\cdots\!\rightarrow\!z_0$. At each step, the standard DDIM update first predicts a clean latent $\hat{z}_0$ and the next trajectory state $z_{t-1}$. TIGA then augments this step with three tightly coupled modules: (i) a \emph{Surrogate-Guided Prior} (SGP) module that aggregates white-box surrogate gradients at $\hat{z}_0$ into a transferable prior direction $\hat{g}_{\text{dir}}$; (ii) a \emph{Surrogate-Guided Directional Search} (SGDS) module that, guided by $\hat{g}_{\text{dir}}$, queries the black-box detector with symmetric finite differences and accumulates a stable momentum direction $m_t$; and (iii) a \emph{Schedule-Aware Trajectory Injection} (SATI) module that scales $m_t$ by the noise schedule, reshapes it in the frequency domain, and injects the result into the trajectory state $z_{t-1}$. The adversarial property therefore emerges as part of generation rather than being attached afterwards.

\textbf{Notation.} Let $\mathcal{D}(\cdot)$ denote the VAE decoder and $\mathcal{T}(\cdot)$ denote the differentiable detector preprocessing (resizing and normalization). We write the composite latent-to-detector mapping as $\mathcal{G}=\mathcal{T}\circ\mathcal{D}$. Following the DDIM formulation, the clean latent predicted at step $t$ is
\begin{equation}
\hat{z}_0=\frac{z_t-\sqrt{1-\alpha_t}\,\epsilon_\theta(z_t,t)}{\sqrt{\alpha_t}},
\label{eq:predx0}
\end{equation}
where $\epsilon_\theta$ is the frozen noise predictor, and $\alpha_t$ is the cumulative noise coefficient. The next trajectory state is obtained by the standard DDIM step, i.e.,
\begin{equation}
z_{t-1}=\sqrt{\alpha_{t-1}}\,\hat{z}_0+\sqrt{1-\alpha_{t-1}-\sigma_t^2}\,\epsilon_\theta+\sigma_t\epsilon,
\label{eq:ddimstep}
\end{equation}
with $\sigma_t$ the DDIM stochasticity level, $\epsilon\!\sim\!\mathcal{N}(0,I)$, and $I$ the identity matrix.

\textbf{Threat model.} We consider a realistic deployment setting in which the diffusion model is frozen and the attacker controls only the sampling process. The target AIGC detector is accessed in a strict black-box manner: only its output confidence can be queried, while its architecture, parameters, and gradients are unavailable. We denote the black-box target by its real-probability response $D_b(x)\!\in\!(0,1)$, where $D_b(x)>0.5$ indicates that $x$ is judged \emph{real}. In addition, the attacker holds $M$ white-box surrogate detectors $\{f_j\}_{j=1}^{M}$, each producing a real-class logit $s_j^{\text{sur}}(z)=f_j(\mathcal{G}(z))$ from which gradients can be back-propagated. 

We note that surrogate guidance is computed from logits, whereas the black-box objective is defined on the sigmoid probability $D_b$; this distinction is kept explicit throughout. The primary attack objective is to maximize the real-class confidence of the target detector, subject only to the image being produced by the injected DDIM trajectory:
\begin{equation}
\max_{\{\Delta_t\}}\ D_b(x_0),\quad \text{s.t.}\ x_0=\mathcal{D}(z_0),\ z_0=\mathrm{DDIM}_{\{\Delta_t\}}(z_T,c),
\label{eq:objective}
\end{equation}
where $\{\Delta_t\}$ are the per-step trajectory injections defined below. We intentionally keep Eq.~\eqref{eq:objective} as a single-objective formulation rather than attaching explicit fidelity, semantic, or frequency penalties: perceptual fidelity is not optimized as a loss, term but instead is encouraged implicitly by the design of the injection itself, namely the schedule-aware bounded scaling and the frequency-domain reshaping introduced in Section~\ref{sec:sati}.

\subsection{Surrogate-Guided Prior}

The first module constructs a transferable adversarial direction from the white-box surrogate detectors, which serves as an informative prior for the subsequent black-box search. At each denoising step, the search and guidance are performed at the predicted clean latent $\hat{z}_0$ in Eq.~\eqref{eq:predx0}, since the black-box and surrogate detectors only operate on clean images: the decoded preview $\mathcal{D}(\hat{z}_0)$ approximates the final synthesized image, whereas the noisy state $z_t$ has no meaningful detector response.

Given the $M$ surrogate detectors, we define their ensemble real-class score in the latent space as
\begin{equation}
\bar{s}^{\text{sur}}(z)=\frac{1}{M}\sum_{j=1}^{M}s_j^{\text{sur}}(z)
=\frac{1}{M}\sum_{j=1}^{M}f_j\big(\mathcal{G}(z)\big),
\label{eq:ensemble}
\end{equation}
and obtain the surrogate-guided prior direction by back-propagating the ensemble score to the latent and normalizing it:
\begin{equation}
\hat{g}_{\text{dir}}=\mathrm{normalize}\!\left(\nabla_{z}\,\bar{s}^{\text{sur}}(z)\Big|_{z=\hat{z}_0}\right).
\label{eq:gdir}
\end{equation}
The prior $\hat{g}_{\text{dir}}$ has two desirable properties. First, it is \emph{transferable}: by aggregating gradients across heterogeneous surrogate architectures, it captures detector-agnostic forensic directions rather than overfitting to a single model. Second, it is \emph{sign-aware}: since $\hat{g}_{\text{dir}}$ is the ascent direction of the real-class score, it already points toward the detector-evasive region, so the subsequent search requires neither a reference sample nor an explicit sign-calibration step to start stably. To balance guidance quality against the cost of back-propagation, the prior is refreshed every $\tau_g$ steps and reused in-between. Notably, $\hat{g}_{\text{dir}}$ only specifies a search \emph{orientation}, as opposed to the attack itself: as the surrogates differ from the target, directly injecting it reduces to a pure transfer attack, and the actual detector-evasive direction is determined by the black-box response in the next module.

\subsection{Surrogate-Guided Directional Search}
Given the prior $\hat{g}_{\text{dir}}$, the second module estimates a reliable trajectory guidance direction by querying the black-box target. Instead of performing isotropic zero-order exploration, which is notoriously high in variance and unstable in the high-dimensional latent space, we constrain the search to an anisotropic distribution centered around the surrogate prior.

\textbf{Anisotropic guided sampling.} Let $\beta_g\!\in\![0,1]$ be a fusion coefficient controlling how strongly the search is biased toward the prior. To draw a probing direction, we first sample an isotropic Gaussian $r\!\sim\!\mathcal{N}(0,I)$ and extract its unit component orthogonal to the prior,
\begin{equation}
r_\perp=\frac{r-\langle r,\hat{g}_{\text{dir}}\rangle\,\hat{g}_{\text{dir}}}
{\big\|r-\langle r,\hat{g}_{\text{dir}}\rangle\,\hat{g}_{\text{dir}}\big\|},
\label{eq:rperp}
\end{equation}
and then mix the prior with the orthogonal exploration component to form the probing direction
\begin{equation}
\tilde{u}=\mathrm{normalize}\!\big(\beta_g\,\hat{g}_{\text{dir}}+(1-\beta_g)\,r_\perp\big),
\qquad \tilde{u}\sim\mathcal{U}_{\hat{g}}(\beta_g).
\label{eq:guidedist}
\end{equation}
Here, $\mathcal{U}_{\hat{g}}(\beta_g)$ denotes the resulting anisotropic direction distribution: it concentrates probing energy along the transferable prior while retaining orthogonal exploration to correct for surrogate-target mismatch.

\textbf{Symmetric finite-difference response.} For each probing direction $\tilde{u}_k$, we measure how fast the black-box real-probability changes along it using a symmetric (central) finite difference with probing radius $\rho$:
\begin{equation}
\phi_k=\frac{D_b\!\big(\mathcal{D}(\hat{z}_0+\rho\,\tilde{u}_k)\big)-D_b\!\big(\mathcal{D}(\hat{z}_0-\rho\,\tilde{u}_k)\big)}{2\rho}.
\label{eq:response}
\end{equation}
The scalar $\phi_k$ acts as a directional response weight: a large positive $\phi_k$ indicates that moving along $+\tilde{u}_k$ increases the real-probability, while $\phi_k<0$ indicates the opposite. The central difference yields an $O(\rho^2)$ truncation error, giving a more accurate estimate than one-sided differencing at the same query cost.

\textbf{Direction estimation.} The per-step guided direction estimate is the response-weighted expectation of the probing directions over the guided distribution, approximated with $N$ samples:
\begin{equation}
d_t=\mathbb{E}_{\tilde{u}\sim\mathcal{U}_{\hat{g}}(\beta_g)}\!\big[\phi(\hat{z}_0,\tilde{u})\,\tilde{u}\big]
\approx\frac{1}{N}\sum_{k=1}^{N}\phi_k\,\tilde{u}_k.
\label{eq:dt}
\end{equation}
Directions that raise the real-probability dominate the weighted sum, while uninformative directions are suppressed toward zero. In the spirit of prior-guided black-box estimators such as P-RGF and Guided ES~\cite{cheng2019improving,maheswaranathan2019guided}, this per-step estimate interpolates between transfer and query-based search. At the boundary $\beta_g=0$, the probing direction reduces to $\tilde{u}=r_\perp$, i.e., isotropic zero-order exploration restricted to the subspace orthogonal to the prior; in the high-dimensional latent space a random direction is almost surely nearly orthogonal to $\hat{g}_{\text{dir}}$, so this closely approximates fully isotropic search. At the opposite boundary $\beta_g=1$, all probes align with $\hat{g}_{\text{dir}}$ and Eq.~\eqref{eq:dt} reduces to a one-dimensional black-box line search along the surrogate prior, where the direction is fixed to the prior but its sign and magnitude are still determined by the target queries $\phi_k$, rather than to a query-free transfer attack. 

The intermediate regime $\beta_g\!\in\!(0,1)$ constrains the search subspace with the white-box prior, substantially reducing estimation variance under a fixed query budget. Unlike these methods, which estimate a single direction on a fixed image, here the estimate $d_t$ is one link in a trajectory-level chain: it is accumulated into momentum across denoising steps and injected back into the DDIM trajectory through the schedule-aware, frequency-shaped update of the next module, so that the per-step search and the diffusion sampling process are tightly coupled, rather than applied to a static input.

\textbf{Decayed momentum accumulation.} To stabilize the trajectory-level signal across steps, we accumulate the normalized direction estimates using a step-decayed exponential moving average. We index the denoising steps in sampling order by $i=0,1,\dots,S-1$, so that step $i$ corresponds to DDIM time $t=T-i$, and $i=0$ corresponds to the first, highest-noise step. Writing $d_i$ for the per-step estimate $d_t$ of Eq.~\eqref{eq:dt} at that step, we define $\hat{d}_i=\mathrm{normalize}(d_i)$ and set $\beta_i=\beta_{\mathrm{momentum}}\big(1-\tfrac{i}{S-1}\big)$. The momentum is bootstrapped from the first estimate and updated as
\begin{equation}
m_0=\hat{d}_0,\qquad m_i=\beta_i\,m_{i-1}+(1-\beta_i)\,\hat{d}_i.
\label{eq:momentum}
\end{equation}
Since $\beta_i$ is the largest in the early, high-noise steps and decays toward zero as sampling proceeds, the momentum relies more on the accumulated direction early on, where individual estimates are noisier, and becomes more responsive to the current estimate in the later, low-noise steps.

To control the query cost, the black-box directional search is executed only once every $\tau_q$ denoising steps: on a query step the $N$ finite-difference probes are issued and the momentum $m$ is updated as above; on the intervening steps no query is made and the most recent momentum is reused for injection, with no update applied before the first query step. The default configuration uses $\tau_q=1$, i.e., a guided search at every step; a larger $\tau_q$ trades attack strength for fewer queries, and this query interval is the parameter varied in the corresponding ablation.

\begin{algorithm}[t]
\caption{TIGA: Trajectory-Injected Generative Attack}
\label{alg:sgds}
\begin{algorithmic}[1]
\REQUIRE frozen diffusion model, condition $c$, noise $z_T$, black-box $D_b$, surrogates $\{f_j\}$, coefficients $\beta_g,\beta_{\mathrm{momentum}},\lambda_0$, samples $N$, radius $\rho$, prior-refresh interval $\tau_g$, query interval $\tau_q$
\STATE $m \leftarrow \varnothing$
\FOR{$t=T$ \textbf{to} $1$}
  \STATE $\epsilon_\theta \leftarrow \texttt{UNet}(z_t,t,c)$
  \STATE compute $\hat{z}_0$ by Eq.~\eqref{eq:predx0}, $z_{t-1}$ by Eq.~\eqref{eq:ddimstep}
  \IF{$t \bmod \tau_g = 0$}
    \STATE $\hat{g}_{\text{dir}} \leftarrow \mathrm{normalize}(\nabla_z\bar{s}^{\text{sur}}(z)|_{\hat{z}_0})$ \hfill // SGP
  \ENDIF
  \IF{$t \bmod \tau_q = 0$}
    \FOR{$k=1$ \textbf{to} $N$}
      \STATE sample $\tilde{u}_k \sim \mathcal{U}_{\hat{g}}(\beta_g)$ by Eq.~\eqref{eq:guidedist}
      \STATE $\phi_k \leftarrow$ symmetric finite difference by Eq.~\eqref{eq:response}
    \ENDFOR
    \STATE $d_t \leftarrow \frac{1}{N}\sum_k \phi_k \tilde{u}_k$;\ \ update $m$ by Eq.~\eqref{eq:momentum} \hfill // SGDS
  \ENDIF
  \IF{$m \neq \varnothing$}
    \STATE $\Delta_t \leftarrow$ scale and frequency-reshape $m$ by Eqs.~\eqref{eq:inject}--\eqref{eq:freq}
    \STATE $z_{t-1} \leftarrow \mathrm{clip}(z_{t-1}+\Delta_t,-4,4)$ \hfill // SATI
  \ENDIF
\ENDFOR
\RETURN $x_0=\mathcal{D}(z_0)$
\end{algorithmic}
\end{algorithm}

\subsection{Schedule-Aware Trajectory Injection}
\label{sec:sati}

The third module converts the momentum direction $m_t$ into an actual trajectory update. A key design choice is \emph{where} and \emph{how strongly} to inject. We inject into the trajectory state $z_{t-1}$ from Eq.~\eqref{eq:ddimstep} rather than into the clean prediction $\hat{z}_0$: $\hat{z}_0$ is a transient preview that is recomputed and discarded at every step, whereas $z_{t-1}$ is the actual variable propagated along the trajectory. Perturbing $z_{t-1}$ allows the adversarial signal to accumulate across steps and be progressively absorbed by the remaining denoising, which is the essence of trajectory-level injection.

\textbf{Schedule-aware scaling.} TIGA operates under the stochastic DDIM sampler ($\eta=1$ in all experiments), for which the stochasticity level $\sigma_t$ in Eq.~\eqref{eq:ddimstep} is strictly positive and decays monotonically along the denoising process. Since the accumulated momentum $m_t$ ($\equiv m_i$ at the corresponding step) is a unit direction, we scale it by this stochasticity level $\sigma_t$ and a strength coefficient $\lambda_0$, with $n$ the latent dimension, and clip the magnitude to bound per-element distortion:
\begin{equation}
\Delta_t=\sigma_t\,\lambda_0\,m_t\sqrt{n},\qquad
\Delta_t\leftarrow\mathrm{clip}\!\left(\Delta_t,-\tfrac{\lambda_0}{2},\tfrac{\lambda_0}{2}\right).
\label{eq:inject}
\end{equation}
Because $\sigma_t$ decays monotonically along the denoising process, the injection is strong in early high-noise steps, where it can reshape the trajectory toward the detector-evasive region, and vanishes in late low-noise steps, where the diffusion model recovers clean high-frequency details. This schedule-aware coupling ties the injection to the sampler's own stochasticity so that the perturbation is naturally assimilated rather than appearing as an external noise layer.

\textbf{Frequency-domain artifact suppression.} To further improve fidelity and weaken detector-sensitive forensic traces, we reshape $\Delta_t$ in the frequency domain. We attenuate its high-frequency component via low-pass energy preservation (with $\mathrm{Down}/\mathrm{Up}$ denoting bilinear resampling), and remove the per-channel spatial mean to eliminate the zero-frequency bias that would otherwise cause color shift or over-exposure:
\begin{align}
\Delta_t^{\text{low}}&=\mathrm{Up}\big(\mathrm{Down}(\Delta_t)\big),\nonumber\\
\Delta_t&\leftarrow\Delta_t^{\text{low}}+\tfrac{1}{2}\big(\Delta_t-\Delta_t^{\text{low}}\big),\label{eq:freq}\\
\Delta_t&\leftarrow\Delta_t-\overline{\Delta_t}^{(H,W)}.\nonumber
\end{align}
Attenuating the high-frequency residual suppresses visible texture artifacts and, at the same time, reduces the abnormal spectral fingerprints that frequency-domain forensic detectors rely on. Finally, the reshaped update is injected and the trajectory state is clamped to the valid VAE range:
\begin{equation}
z_{t-1}\leftarrow\mathrm{clip}\big(z_{t-1}+\Delta_t,\,-4,\,4\big).
\label{eq:final}
\end{equation}
The updated $z_{t-1}$ becomes the input of the next denoising step, propagating the adversarial signal forward until the final detector-evasive image $x_0=\mathcal{D}(z_0)$ is obtained. The complete procedure, with the three modules embedded inside the standard DDIM denoising loop over a frozen diffusion model, is summarized in Algorithm~\ref{alg:sgds}.

\section{Experiments}

\begin{table*}[!t]
  \centering
  \caption{Black-box attack performance and visual quality under four black-box target settings on REDS (R: ResNet-50, E: EfficientNet-B0, D: DeiT, S: Swin-T). ASR$\uparrow$ denotes the attack success rate (higher is stronger); BRIS$\downarrow$ denotes the BRISQUE no-reference quality score (lower is better). The best result in each column is shown in \textbf{bold}, and our method is highlighted in gray.}
  \label{tab:blackbox_attack}

  \renewcommand{\arraystretch}{1.}
    \scalebox{1.10}{
  \begin{tabular}{c cc cc cc cc cc}
  \toprule
  \multirow{2}{*}{\textbf{Attacks}}
  & \multicolumn{2}{c}{\textbf{ResNet-50}} & \multicolumn{2}{c}{\textbf{EfficientNet-B0}}
  & \multicolumn{2}{c}{\textbf{DeiT}} & \multicolumn{2}{c}{\textbf{Swin-T}}
  & \multicolumn{2}{c}{\textbf{Average}} \\
  \cmidrule(lr){2-3}\cmidrule(lr){4-5}\cmidrule(lr){6-7}\cmidrule(lr){8-9}\cmidrule(lr){10-11}
  & ASR$\uparrow$ & BRIS$\downarrow$ & ASR$\uparrow$ & BRIS$\downarrow$
  & ASR$\uparrow$ & BRIS$\downarrow$ & ASR$\uparrow$ & BRIS$\downarrow$
  & ASR$\uparrow$ & BRIS$\downarrow$ \\
  \midrule
  FGSM   & 86.64 & 87.37 & 99.92 & 106.63 & 91.13 & 106.63 & 96.37 & 106.63 & 93.52 & 101.82 \\
  PGD    & 28.15 & 36.43 & 83.27 & 36.05 & 84.98 & 36.05 & 94.43 & 36.06 & 72.71 & 36.15 \\
  SimBA  & 78.63 & 26.79 & 84.29 & 27.19 & 68.71 & 28.33 & 51.86 & 27.53 & 70.87 & 27.46 \\
  Square & 99.14 & 69.20 & \textbf{100.00} & 78.11 & \textbf{100.00} & 57.64 & 99.86 & 52.39 & 99.89 & 64.34 \\
  BruSLe & 99.11 & 35.40 & \textbf{100.00} & 35.05 & \textbf{100.00} & 35.14 & 89.14 & 35.93 & 97.06 & 35.38 \\
  R$^2$BA& \textbf{100.00} & 21.76 & \textbf{100.00} & 27.39 & \textbf{100.00} & 25.83 & 98.43 & 35.93 & 99.61 & 24.89 \\
  \rowcolor{gray!12}
  \textbf{Ours} & \textbf{100.00} & \textbf{20.43} & \textbf{100.00} & \textbf{21.28} & \textbf{100.00} & \textbf{19.18} & \textbf{100.00} & \textbf{20.05} & \textbf{100.00} & \textbf{20.24} \\
  \bottomrule
  \end{tabular}}
  \end{table*}

\begin{table*}[!t]
  \centering
  \caption{Transfer-attack evasion rate (\%, percentage judged ``real'') against five unseen detectors (CNN, DIRE, Uni, Effort, PGC), none used as a white-box surrogate or black-box target during optimization, evaluated separately for each of the four REDS black-box target settings used to craft the images. ``Clean'': unattacked images. Bold: best per column; gray row: our method.}
  \label{tab:transfer_attack}
  
  \renewcommand{\arraystretch}{1.1}
  \setlength{\tabcolsep}{8pt}
  \scalebox{1.1}{
  \begin{tabular}{cccccc@{\hspace{12pt}}cccccc}
  \toprule
  & \multicolumn{5}{c}{\textbf{Target Detector: DeiT}}
  & & \multicolumn{5}{c}{\textbf{Target Detector: EfficientNet-B0}} \\
  \cmidrule(lr){2-6}\cmidrule(lr){8-12}
 \textbf{Methods} & \textbf{CNN} & \textbf{DIRE} & \textbf{Uni} & \textbf{Effort} & \textbf{PGC}
  &
  \textbf{Methods} & \textbf{CNN} & \textbf{DIRE} & \textbf{Uni} & \textbf{Effort} & \textbf{PGC} \\
  \midrule
  Clean  & 77.86 & 90.57 & 21.29 & 22.14 & 4.29
  &
  Clean  & 77.86 & 90.57 & 21.29 & 22.14 & 4.29 \\

  FGSM   & \textbf{100.00} & 82.86 & 35.14 & 0.86  & 7.86
  &
  FGSM   & \textbf{100.00} & 82.86 & 35.14 & 0.86  & 7.86 \\

  PGD    & \textbf{100.00} & 83.43 & 5.57  & 0.00  & 0.00
  &

  PGD    & \textbf{100.00} & 83.43 & 5.57  & 0.00  & 0.00 \\

  SimBA  & 98.57  & 89.97 & 14.86 & 4.43  & 1.57
  &
  SimBA  & 99.86  & 89.72 & 11.43 & 1.86  & 1.00 \\

  Square & \textbf{100.00} & 92.43 & 6.86  & 0.71  & 0.86
  &
  Square & \textbf{100.00} & 92.43 & 9.43  & 1.14  & 1.14 \\

  BruSLe & \textbf{100.00} & 95.86 & 33.58 & 56.86 & 0.43
  &
  BruSLe & \textbf{100.0} & 96.43 & 31.71 & 54.14 & 0.43 \\

  R$^2$BA & \textbf{100.00} & 95.72 & 20.86 & 15.86 & 9.29
  &
  R$^2$BA & 98.71 & 94.28 & 18.14 & 11.43 & 12.57 \\ 

  \rowcolor{gray!12}
  \textbf{Ours} & 99.53 & \textbf{100.00} & \textbf{68.40} & \textbf{86.32} & \textbf{49.06}
  &
  \textbf{Ours} & \textbf{100.00} & \textbf{99.53} & \textbf{47.17} & \textbf{66.98} & \textbf{22.17} \\
  \midrule

  & \multicolumn{5}{c}{\textbf{Target Detector: ResNet-50}}
  & & \multicolumn{5}{c}{\textbf{Target Detector: Swin-T}} \\
  \cmidrule(lr){2-6}\cmidrule(lr){8-12}
  \textbf{Methods} & \textbf{CNN} & \textbf{DIRE} & \textbf{Uni} & \textbf{Effort} & \textbf{PGC}
  &
  \textbf{Methods} & \textbf{CNN} & \textbf{DIRE} & \textbf{Uni} & \textbf{Effort} & \textbf{PGC} \\
  \midrule

  Clean  & 77.86 & 90.57 & 21.29 & 22.14 & 4.29
  &
  Clean  & 77.86 & 90.57 & 21.29 & 22.14 & 4.29 \\

  FGSM   & \textbf{100.00} & 84.79 & 41.14 & 7.71  & 5.00
  &
  FGSM   & \textbf{100.00} & 82.86 & 35.14 & 0.86  & 7.86 \\

  PGD    & \textbf{100.00} & 84.15 & 15.43 & 0.00  & 0.57
  &
  PGD    & \textbf{100.00} & 83.43 & 5.57  & 0.00  & 0.00 \\

  SimBA  & 98.71  & 90.29 & 11.14 & 6.29  & 0.86
  &
  SimBA  & 99.71  & 89.86 & 12.29 & 2.57  & 0.86 \\

  Square & \textbf{100.00} & 92.58 & 7.86  & 0.71  & 0.86
  &
  Square & \textbf{100.00} & 92.15 & 7.43  & 0.14  & 0.86 \\

  BruSLe & \textbf{100.00} & 95.67 & 31.43 & 55.71 & 0.43
  &
  BruSLe & \textbf{100.00} & 97.86 & 34.29 & 59.71 & 0.43 \\

  R$^2$BA & \textbf{100.00} & 95.29 & 19.86 & 21.86 & 12.86
  &
  R$^2$BA & \textbf{100.00} & 94.65 & 22.14 & 35.57 & 12.73 \\ 

  \rowcolor{gray!12}
  \textbf{Ours} & \textbf{100.00} & \textbf{100.00} & \textbf{62.74} & \textbf{75.94} & \textbf{40.62}
  &
  \textbf{Ours} & \textbf{100.00} & \textbf{99.53} & \textbf{59.43} & \textbf{77.36} & \textbf{38.21} \\

  \bottomrule
  \end{tabular}}
  \end{table*}

\subsection{Experimental Setup}
\label{sec:setup}

\textbf{Implementation details.} TIGA is implemented in PyTorch~1.7 with CUDA~11.0 on top of a frozen, pretrained Collaborative Diffusion backbone, and all experiments are run on an NVIDIA RTX~3090 GPU. % TODO(user): confirm the exact GPU model used for the reported experiments.
The backbone synthesizes $512\times512$ face images conditioned jointly on a segmentation mask and a free-form text description, operating in a $3\times64\times64$ latent space; for every test instance, we pair a $20$-class CelebAMask-HQ-style segmentation mask~\cite{CelebAMask-HQ} (resized to $32\times32$ and one-hot encoded) with a textual attribute description (e.g., age, hairstyle, or beard length) to form the condition $c=(\text{mask},\text{text})$ used throughout Section~\ref{sec:method}. DDIM sampling runs for $S=50$ steps with stochasticity $\eta=1$, and for each instance the reference (unattacked) and adversarial trajectories share the same initial noise $z_T$ and condition $c$, so that any difference between them is attributable only to the trajectory injection rather than to sampling variance. Unless otherwise noted, every reported TIGA result uses the full configuration that also serves as the reference point of the ablation study (Section~\ref{sec:ablation}): guidance strength $\lambda_0=0.35$, surrogate-prior fusion coefficient $\beta_{\mathrm{guide}}=0.5$, $N=10$ probing directions per guided step with finite-difference radius $\rho=0.05$, query interval $\tau_q=1$, so that guided search is performed at \emph{every} denoising step, momentum decay $\beta_{\mathrm{momentum}}=0.8$, and surrogate-prior refresh interval $\tau_g=5$. Images are generated in batches under a fixed random seed so that results are reproducible across methods.

\textbf{Generic AIGC detectors.} We use four architecturally diverse, binary real/fake AIGC detectors as the target/surrogate pool, jointly denoted \textbf{REDS}: ResNet-50 (\textbf{R}, CNN)~\cite{he2016deep}, EfficientNet-B0 (\textbf{E}, CNN)~\cite{tan2019efficientnet}, DeiT-Base (\textbf{D}, plain Vision Transformer)~\cite{touvron2021training}, and Swin-Base (\textbf{S}, hierarchical Transformer)~\cite{liu2021swin}. Each detector is fine-tuned in-domain on $12{,}000$ images synthesized by our diffusion backbone (balanced real/fake) as a single real/fake logit ($\mathrm{sigmoid}(\cdot)>0.5$ indicates ``real''), and all four reach over $95\%$ accuracy on a held-out test set, so that the evaluated targets are reliable rather than weak detectors. Following the threat model in Section~\ref{sec:method}, we construct four black-box settings by cycling through REDS: in each setting, one model plays the role of the strict black-box target $D_b$, queried only through its output probability, while the remaining three serve as the white-box surrogates $\{f_j\}_{j=1}^{3}$ used by the SGP module. All four settings share the same test set of (mask, text) pairs.

\textbf{Transfer detectors.} To measure whether the induced evasiveness generalizes beyond the REDS pool used during optimization, we additionally consider five specialized forensic detectors that are never used as either a white-box surrogate or a black-box target: a CNN-based spatial-artifact classifier (\textbf{CNN})~\cite{wang2020cnn}, a diffusion-reconstruction-error detector (\textbf{DIRE})~\cite{wang2023dire}, a CLIP-feature-based universal detector (\textbf{Uni})~\cite{ojha2023towards}, and two further generalizable detectors (\textbf{Effort}~\cite{pmlr-v267-yan25b} and \textbf{PGC})~\cite{zhou2026pgc}. All five are trained on $144$K images generated by Stable Diffusion v1.4~\cite{rombach2022high}  and ProGAN covering the same four object categories, with the ProGAN data drawn from ForenSynths~\cite{wang2020cnn} and the SD-v1.4 data from GenImage~\cite{zhu2023genimage}. We report their evasion behavior in Section~\ref{sec:comparison}, and use the abbreviations CNN/DIRE/Uni/Effort/PGC in all tables.

\textbf{Baselines.} We compare TIGA against six representative adversarial attacks that are adapted to AIGC-detector evasion: two white-box gradient attacks, FGSM~\cite{goodfellow2014explaining} and PGD~\cite{madry2018towards}; three query-based black-box attacks, SimBA~\cite{guo2019simple}, Square Attack~\cite{andriushchenko2020square}, and BruSLe~\cite{vo2024brusleattack}; and the most recent AIGC-oriented black-box attack, R$^2$BA~\cite{xie2025take}, which optimizes over realistic post-processing operations. All baselines are applied as \emph{post-hoc} pixel-space perturbations on top of the identical clean image produced by the same DDIM trajectory as TIGA, sharing the same initial noise $z_T$, mask, and text, so that the underlying generative content is held fixed across methods and only the perturbation mechanisms differ. For fair comparisons, all attacks share the same maximum perturbation budget $\epsilon=16/255$. The four query-based black-box attacks, SimBA, Square, BruSLe, and R$^2$BA, each query the black-box target under a fixed budget of $1{,}000$ queries with at most $20$ iterations, are launched from a random initialization, and never access the target gradients. As FGSM and PGD require gradients that are unavailable under the black-box target setting, we run them as transfer attacks: their perturbations are crafted on a single white-box surrogate detector and then transferred to the black-box target. Neither method issues any query to the target or accesses its gradients.

\textbf{Evaluation metrics.} We report: (i) the Attack Success Rate (ASR$\uparrow$), the percentage of images that the target detector correctly labels ``fake'' before the attack and flips to ``real'' afterward; (ii) Blind/Referenceless Image Spatial Quality Evaluator (BRISQUE$\downarrow$), a no-reference perceptual quality score computed on the final adversarial image, where a lower value indicates fewer visible artifacts; (iii) the \textbf{detector-evasion rate} on fully unseen detectors, used to quantify transferability; and (iv) \textbf{robust ASR} under common post-processing operations.

\subsection{Comparison Study}
\label{sec:comparison}

\begin{figure*}
\centering
\includegraphics[width=0.92\linewidth]{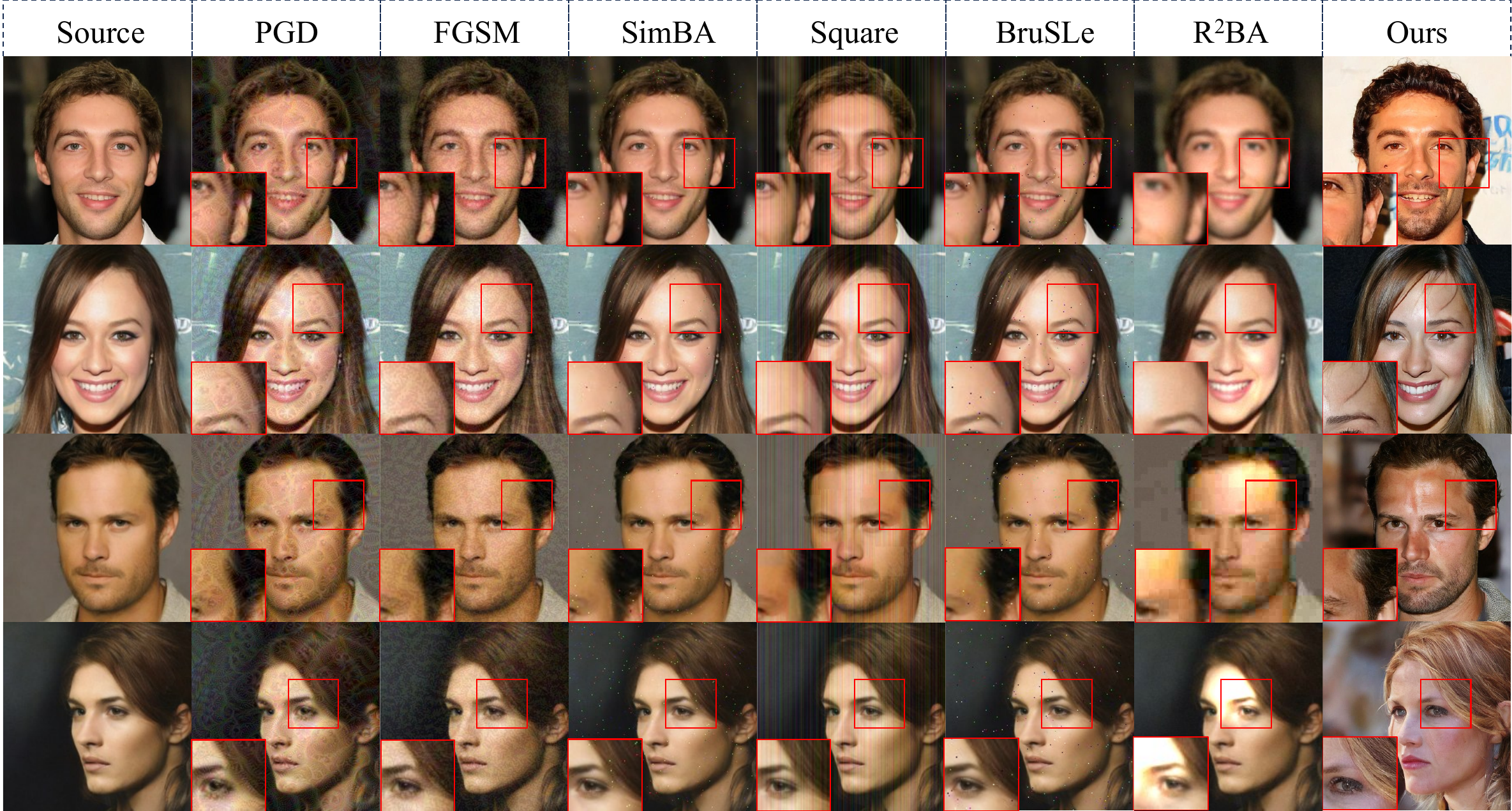}
\caption{Qualitative comparison of adversarial faces produced by each attack from the same DDIM-generated source image (leftmost column); red boxes zoom into the forehead, cheek, and periocular skin.}
\label{fig:compare_visual}
\end{figure*}

\subsubsection{Attack Effectiveness and Visual Quality on REDS}

Table~\ref{tab:blackbox_attack} compares TIGA against the six baselines on all four REDS targets. TIGA attains a perfect $100\%$ ASR on every target while obtaining the lowest BRISQUE in every column (average $20.24$), i.e., it Pareto-dominates every baseline on both axes at once. Among the baselines, R$^2$BA is the strongest overall, nearly saturating ASR (average $99.61\%$) at the best perceptual quality of any baseline (average BRISQUE $24.89$), yet it still trails TIGA on BRISQUE at every target. Square and BruSLe also saturate ASR (average $99.89\%$/$97.06\%$) but at a much higher visual cost (BRISQUE $64.34$/$35.38$), SimBA is moderate in quality ($27.46$) but weak in ASR ($70.87\%$), and the gradient attacks are the least consistent (FGSM: $93.52\%$ ASR at the worst BRISQUE $101.82$; PGD collapses to $28.15\%$ on ResNet-50). Once a perturbation is externally superimposed in pixel space, an attacker is forced to trade success rate against visible artifacts; TIGA, by shaping the sample \emph{during} generation rather than after it, avoids this trade-off and dominates all six baselines on both axes.

\subsubsection{Qualitative Comparison}

\begin{table*}[t]
\centering
\caption{ASR$\uparrow$ (\%) of seven attack methods under different
post-processing operations and parameter settings. For each target detector
and each post-processing setting, the \textbf{bold} value denotes the highest
ASR among all attack methods. Gray row denotes our method.}
\label{tab:robustness}
\scriptsize
\setlength{\tabcolsep}{2.2pt}
\renewcommand{\arraystretch}{1.08}

\resizebox{\textwidth}{!}{%
\begin{tabular}{ccccccc ccccc ccccc ccccc}
\toprule
\multirow{2}{*}{\textbf{Detector}}
& \multirow{2}{*}{\textbf{Attack}}
& \multicolumn{5}{c}{\textbf{Gaussian Blur}}
& \multicolumn{5}{c}{\textbf{Gaussian Noise}}
& \multicolumn{5}{c}{\textbf{JPEG Compression}}
& \multicolumn{5}{c}{\textbf{Resize}} \\
\cmidrule(lr){3-7}
\cmidrule(lr){8-12}
\cmidrule(lr){13-17}
\cmidrule(lr){18-22}

& & 0.5 & 1 & 1.5 & 2 & 3
& 0.01 & 0.03 & 0.05 & 0.08 & 0.1
& 10 & 30 & 50 & 75 & 95
& 0.25 & 0.5 & 0.75 & 1.25 & 1.5 \\
\midrule

% ============================================================
% R
% ============================================================
\multirow[c]{6}{*}{\textbf{R}}
& FGSM
& 86.18 & 90.54 & 87.57 & 73.86 & 44.35
& 85.57 & 85.43 & 92.62 & 99.43 & \textbf{100.00}
& 93.71 & 84.14 & 85.71 & 84.43 & 85.79
& 38.29 & 87.43 & 89.86 & 87.29 & 89.97 \\

& PGD
& 37.14 & 42.57 & 37.29 & 25.14 & 15.23
& 27.18 & 33.29 & 62.71 & 97.86 & \textbf{100.00}
& 88.43 & 29.43 & 31.29 & 29.14 & 27.14
& 6.14 & 35.43 & 40.87 & 36.43 & 38.43 \\

& Square
& 43.86 & 40.28 & 40.43 & 32.43 & 25.86
& 60.29 & 56.57 & 77.57 & 98.57 & \textbf{100.00}
& 93.43 & 69.57 & 86.57 & 87.86 & 82.43
& 9.14 & 31.29 & 43.71 & 54.57 & 49.43 \\

& SimBA
& 2.86 & 5.57 & 7.14 & 6.86 & 9.14
& 0.86 & 8.86 & 41.86 & 94.71 & \textbf{100.00}
& 86.00 & 2.43 & 1.29 & 1.00 & 1.86
& 1.57 & 2.71 & 3.86 & 5.17 & 4.86 \\

& BrusLe
& 89.86 & 41.29 & 17.29 & 11.14 & 8.29
& 97.43 & 80.71 & 73.43 & 96.71 & 99.86
& 94.52 & 67.86 & 77.42 & 86.86 & 97.71
& 3.96 & 22.29 & 61.33 & 83.71 & 83.71 \\

& R$^2$BA
& 92.86 & 85.43 & 82.71 & 73.14 & 52.67
& 95.14 & 87.52 & 91.72 & 97.29 & \textbf{100.00}
& \textbf{98.86} & 61.86 & 50.14 & 67.43 & 94.29
& 33.43 & 77.68 & 82.43 & 84.17 & 85.14 \\

\rowcolor{gray!12}
& Ours
& \textbf{94.35} & \textbf{93.82} & \textbf{91.46}
& \textbf{84.73} & \textbf{73.68}
& \textbf{98.72} & \textbf{92.14} & \textbf{95.63}
& \textbf{99.71} & \textbf{100.00}
& 97.29 & \textbf{90.64} & \textbf{92.37}
& \textbf{93.18} & \textbf{98.52}
& \textbf{78.42} & \textbf{91.68} & \textbf{94.15}
& \textbf{92.83} & \textbf{94.36} \\
\midrule

% ============================================================
% E
% ============================================================
\multirow[c]{6}{*}{\textbf{E}}
& FGSM
& \textbf{100.00} & \textbf{100.00} & 99.71 & 98.29 & 76.87
& \textbf{100.00} & \textbf{100.00} & \textbf{100.00}
& \textbf{100.00} & \textbf{100.00}
& 98.14 & 99.71 & 99.86 & 99.43 & \textbf{100.00}
& 85.14 & 99.43 & \textbf{100.00} & \textbf{100.00}
& \textbf{100.00} \\

& PGD
& \textbf{100.00} & \textbf{100.00} & 98.29 & 90.71 & 52.86
& \textbf{100.00} & 99.86 & 99.23 & 99.86 & \textbf{100.00}
& 82.86 & 98.57 & 99.86 & 99.86 & \textbf{100.00}
& 65.43 & 99.12 & \textbf{100.00} & \textbf{100.00}
& \textbf{100.00} \\

& Square
& 60.71 & 30.00 & 35.71 & 25.43 & 15.00
& 78.71 & 82.14 & 91.43 & 99.00 & 99.86
& 60.00 & 58.71 & 77.29 & 73.29 & 83.00
& 7.29 & 31.00 & 47.86 & 58.43 & 58.43 \\

& SimBA
& 11.71 & 5.43 & 6.43 & 6.43 & 7.43
& 22.43 & 49.17 & 83.86 & 98.29 & 99.86
& 31.14 & 3.43 & 1.71 & 1.00 & 12.43
& 3.00 & 1.29 & 3.57 & 7.29 & 9.57 \\

& BrusLe
& 52.43 & 50.86 & 21.57 & 14.86 & 9.71
& 97.43 & 97.43 & 98.43 & 99.86 & \textbf{100.00}
& 54.14 & 58.57 & 80.21 & 88.86 & 82.86
& 4.29 & 19.86 & 53.29 & 60.86 & 65.29 \\

& R$^2$BA
& 89.86 & 30.14 & 27.86 & 23.57 & 22.32
& 92.57 & 97.29 & 99.14 & 99.86 & \textbf{100.00}
& 68.43 & 19.29 & 14.71 & 16.47 & 49.57
& 9.43 & 23.22 & 34.86 & 47.79 & 42.14 \\

\rowcolor{gray!12}
& Ours
& \textbf{100.00} & \textbf{100.00} & \textbf{99.86}
& \textbf{99.14} & \textbf{83.46}
& \textbf{100.00} & \textbf{100.00} & \textbf{100.00}
& \textbf{100.00} & \textbf{100.00}
& \textbf{99.26} & \textbf{99.86} & \textbf{99.93}
& \textbf{99.94} & \textbf{100.00}
& \textbf{90.72} & \textbf{99.71} & \textbf{100.00}
& \textbf{100.00} & \textbf{100.00} \\
\midrule

% ============================================================
% D
% ============================================================
\multirow[c]{6}{*}{\textbf{D}}
& FGSM
& 90.17 & 87.14 & 83.14 & 71.92 & 45.86
& 91.14 & 92.29 & 94.14 & 95.57 & 96.57
& 92.29 & 91.65 & 89.86 & 91.14 & 91.57
& 60.86 & 85.14 & 89.12 & 90.63 & 90.88 \\

& PGD
& 84.29 & 78.57 & 73.29 & 58.43 & 35.29
& 85.43 & 83.86 & 84.29 & 90.86 & 93.29
& 87.14 & 77.71 & 79.86 & 83.86 & 86.43
& 45.29 & 74.71 & 79.86 & 81.29 & 80.57 \\

& Square
& 79.43 & 74.86 & 65.57 & 46.14 & 29.29
& 75.71 & 71.28 & 76.71 & 85.57 & 91.14
& 82.71 & 54.86 & 64.71 & 65.86 & 74.00
& 32.29 & 66.29 & 76.71 & 78.86 & 76.14 \\

& SimBA
& 30.43 & 27.57 & 23.71 & 12.86 & 10.86
& 27.00 & 37.71 & 56.86 & 77.14 & 85.57
& 77.00 & 22.00 & 18.86 & 19.14 & 26.14
& 9.86 & 20.57 & 25.00 & 24.43 & 22.57 \\

& BrusLe
& 55.86 & 38.71 & 28.71 & 14.57 & 8.14
& 70.29 & 63.43 & 71.57 & 83.57 & 88.86
& 79.29 & 38.43 & 40.14 & 43.79 & 60.71
& 9.12 & 31.14 & 41.29 & 48.57 & 47.14 \\

& R$^2$BA
& 91.14 & 74.29 & 66.17 & 48.43 & 33.65
& 93.43 & 93.57 & 95.75 & 95.26 & 95.19
& 96.71 & 84.52 & 83.86 & 90.57 & 96.29
& 35.86 & 64.31 & 77.39 & 84.95 & 83.57 \\

\rowcolor{gray!12}
& Ours
& \textbf{93.74} & \textbf{91.53} & \textbf{88.96}
& \textbf{82.35} & \textbf{72.84}
& \textbf{94.21} & \textbf{95.18} & \textbf{96.32}
& \textbf{97.46} & \textbf{98.37}
& \textbf{95.46} & \textbf{94.38} & \textbf{93.21}
& \textbf{94.63} & \textbf{97.28}
& \textbf{76.49} & \textbf{89.82} & \textbf{92.47}
& \textbf{93.58} & \textbf{94.11} \\
\midrule

% ============================================================
% S
% ============================================================
\multirow[c]{6}{*}{\textbf{S}}
& FGSM
& 96.43 & 97.39 & 95.57 & 91.26 & 64.29
& 96.29 & 93.86 & 91.29 & 87.86 & 86.43
& 84.71 & 98.71 & 97.86 & 97.43 & 97.43
& 83.21 & 96.92 & 98.29 & 98.67 & 98.67 \\

& PGD
& 93.57 & 92.57 & 90.14 & 80.57 & 48.69
& 90.43 & 75.29 & 64.43 & 62.57 & 68.57
& 64.57 & 88.86 & 90.29 & 90.57 & 96.14
& 68.86 & 90.14 & 94.57 & 94.86 & 95.86 \\

& Square
& 90.14 & 93.14 & 87.57 & 68.29 & 40.29
& 14.86 & 24.71 & 27.43 & 34.29 & 40.71
& 54.71 & 71.14 & 81.71 & 88.57 & 88.86
& 44.86 & 87.71 & 96.86 & 94.29 & 92.57 \\

& SimBA
& 7.00 & 7.00 & 9.29 & 7.86 & 8.43
& 0.14 & 3.71 & 9.86 & 21.57 & 31.86
& 35.14 & 9.71 & 9.43 & 9.71 & 5.43
& 8.14 & 7.57 & 11.71 & 6.71 & 6.29 \\

& BrusLe
& 94.57 & 97.86 & 83.26 & 37.29 & 11.71
& 9.12 & 53.71 & 62.43 & 65.43 & 67.29
& 62.57 & 72.57 & 84.14 & 87.29 & 97.11
& 18.71 & 94.25 & 99.15 & 97.71 & 97.57 \\

& R$^2$BA
& 76.15 & 82.43 & 83.57 & 79.29 & 64.93
& 89.31 & 90.43 & 90.29 & 89.57 & 89.86
& 89.91 & 88.71 & 97.43 & 95.14 & 70.96
& 81.43 & 84.14 & 84.14 & 80.29 & 80.71 \\

\rowcolor{gray!12}
& Ours
& \textbf{97.82} & \textbf{98.64} & \textbf{97.16}
& \textbf{93.87} & \textbf{74.56}
& \textbf{97.38} & \textbf{95.27} & \textbf{93.14}
& \textbf{90.72} & \textbf{93.36}
& \textbf{88.94} & \textbf{99.18} & \textbf{98.42}
& \textbf{98.11} & \textbf{98.36}
& \textbf{97.92} & \textbf{98.16} & \textbf{99.52}
& \textbf{99.14} & \textbf{99.05} \\
\bottomrule
\end{tabular}%
}
\end{table*}

Fig.~\ref{fig:compare_visual} visualizes this trade-off. The six baselines all perturb pixels on top of the identical clean generation, so they preserve the coarse appearance of the source but leave visible signatures in the zoomed patches: uniform speckle noise for FGSM/PGD, sparse local changes for SimBA, colored blotches for Square/BruSLe, and a mild global blur/illumination shift for R$^2$BA. TIGA (rightmost) shows none of these superimposed patterns and keeps skin and hair smooth, consistent with its lowest BRISQUE. This follows from \emph{where} the update is applied: injected into the DDIM trajectory (Eq.~\eqref{eq:inject}) rather than onto a finished image, it is smoothed and regularized by the remaining denoising steps. The trade-off is that, since the schedule-aware injection is the strongest at early, high-noise steps, TIGA samples can show slightly larger coarse, low-frequency changes (e.g., hair color or skin tone) than the pixel-perturbation baselines. This is inherent to the source-image-free setting: TIGA only aims to keep the output a plausible, naturally generated face under the given condition, not pixel-level fidelity to a source; for detector evasion the perceptual naturalness of the final image is the operative criterion.
\subsubsection{Transferability to Unseen Detectors}

To test whether the evasiveness induced by TIGA generalizes beyond the four REDS models used during optimization, we re-score all attacked images with the five unseen transfer detectors described in Section~\ref{sec:setup} (CNN, DIRE, Uni, Effort, and PGC), none of which is used as a white-box surrogate or a black-box target during attack optimization. Any evasion observed reflects genuine cross-architecture, cross-paradigm transfer, rather than optimization against a known model.
% TODO(user): replace universalfakedetect / effort_detector / pgc_detector with the
% actual bibliography keys for Uni / Effort / PGC once finalized.

As reported in Table~\ref{tab:transfer_attack}, two trends stand out. On CNN and DIRE, which respond to low-level, spatially localized statistics, most methods including TIGA reach or nearly reach $100\%$ evasion; the high Clean rate on CNN ($77.86\%$) indicates it is weak even against unattacked samples. More importantly, on the higher-level detectors Uni, Effort, and PGC the baselines stay close to the Clean rate ($21.29\%$/$22.14\%$/$4.29\%$), whereas TIGA improves substantially and consistently across all four settings (Uni: $47.17$--$68.40\%$; Effort: $66.98$--$86.32\%$; PGC: $22.17$--$49.06\%$). Even R$^2$BA, the strongest baseline on REDS, stays near Clean here (e.g., PGC $12.86\%$ vs.\ our $40.62\%$ under the ResNet-50 target). This indicates that the baselines' post-hoc perturbations overfit to low-level statistics of their own target and do not transfer to higher-level detectors, while TIGA's trajectory-level modification shifts properties of the generated content itself and transfers more consistently.

Transfer strength also depends on the surrogate set: the DeiT-target setting (surrogates $\{$R,E,S$\}$) transfers the best to Uni/Effort/PGC ($68.40\%$/$86.32\%$/$49.06\%$) and the EfficientNet-target setting (surrogates $\{$R,D,S$\}$) transfers the worst ($47.17\%$/$66.98\%$/$22.17\%$), even though both share R and S. Since the two differ only in EfficientNet vs.\ DeiT, including a CNN-style surrogate with a different inductive bias from ResNet-50 contributes disproportionately to the prior's transferability.

\subsubsection{Robustness Evaluation}
\label{sec:robustness}

To assess whether the evasiveness of each method survives common image post-processing that an AIGC platform might apply before running detection, we take the adversarial images produced by each method against all four REDS targets and re-score them, after Gaussian blur, Gaussian noise, JPEG compression, or resizing, using all four REDS detectors; Table~\ref{tab:robustness} reports, for each (perturbation family, parameter) setting and each scoring detector, the percentage of processed images still classified as ``real.''

Several observations follow from Table~\ref{tab:robustness}, one per post-processing family. Under \textbf{Gaussian blur}, TIGA degrades gracefully (ASR-R $94.35\%\!\rightarrow\!73.68\%$ from radius $0.5$ to $3$), whereas SimBA collapses to single digits already at radius $0.5$ and Square/BruSLe fall $40$--$60$ points; R$^2$BA is the most blur-robust baseline ($92.86\%\!\rightarrow\!52.67\%$) but still trails TIGA at every level. Under \textbf{JPEG compression}, TIGA stays close to its uncompressed level (e.g., $97.29\%$ at quality $10$), while the baselines are the weakest at intermediate qualities ($30$--$75$): SimBA and PGD fall to $2.43\%$ and $29.43\%$ at quality $30$, and even R$^2$BA drops to $50.14\%$ at quality $50$. Under \textbf{Resize}, at the largest downsampling ($0.25$) SimBA/BruSLe/Square/PGD fall to single digits while TIGA retains $78.42\%$. Under \textbf{Gaussian noise}, TIGA is the strongest at $\sigma\le0.05$, but at $\sigma=0.10$ essentially all methods converge to near $100\%$; we read this as a metric ceiling effect, since the noise itself pushes images out of the detectors' training distribution regardless of any perturbation. Across the three families that remove or attenuate structure (blur, compression, resize), TIGA is the only method whose evasiveness survives realistic, moderate-strength post-processing, consistent with its adversarial signal being a property of the generated content rather than an added high-frequency pattern.

\subsection{Ablation Study}
\label{sec:ablation}

\begin{figure*}[t]
    \centering
    % -------------------- Top: ASR --------------------
    \begin{minipage}[t]{1\textwidth}
        \centering
        \includegraphics[width=\linewidth]{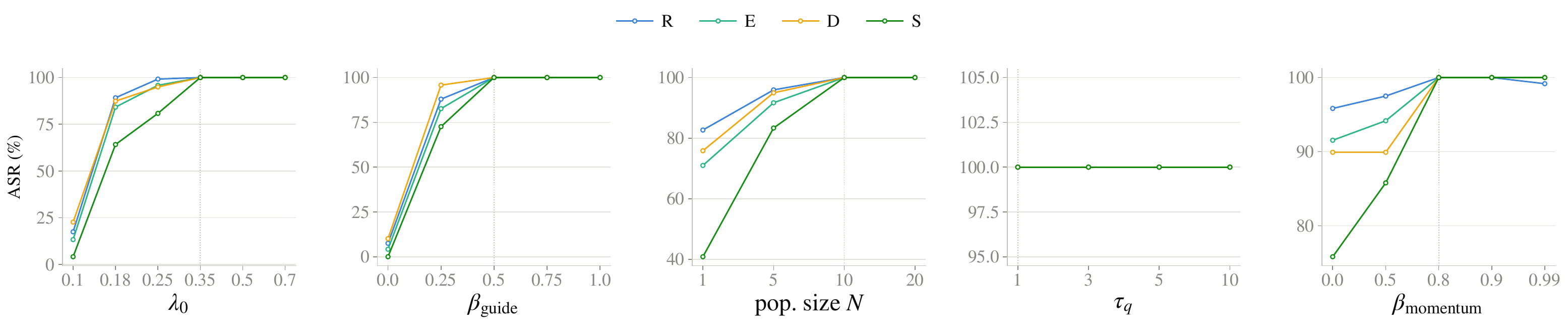}
        \vspace{-2mm}
        \caption*{(a) Attack success rate.}
    \end{minipage}

    \vspace{2mm}

    % -------------------- Bottom: BRISQUE --------------------
    \begin{minipage}[t]{1\textwidth}
        \centering
        \includegraphics[width=\linewidth]{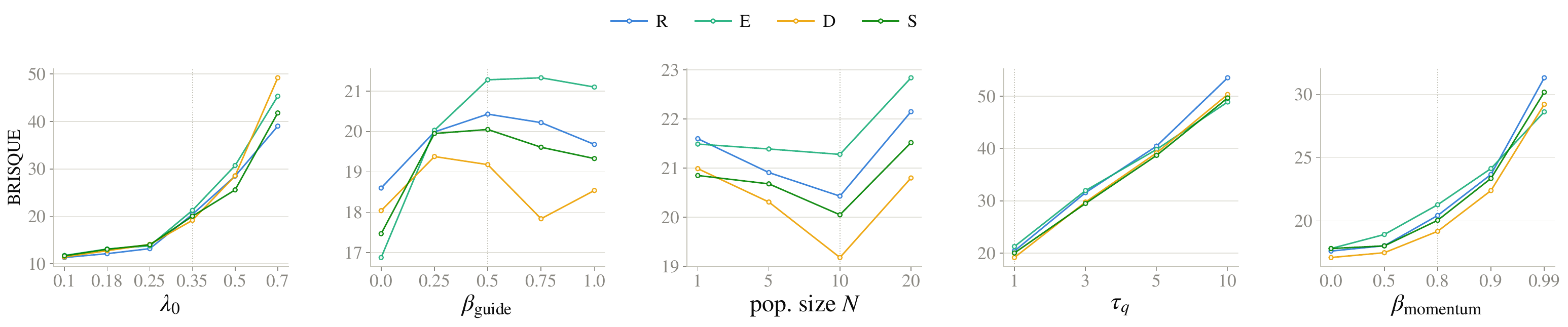}
        \vspace{-2mm}
        \caption*{(b) Perceptual quality measured by BRISQUE.}
    \end{minipage}

    \vspace{-1mm}
    \caption{Sensitivity of (a) ASR and (b) BRISQUE to each hyperparameter
    ($\lambda_0$, $\beta_{\mathrm{guide}}$, population size $N$,
    query interval $\tau_q$, and $\beta_{\mathrm{momentum}}$). Each hyperparameter is
    swept independently while the remaining four are fixed at their
    full-configuration values. Results are evaluated against four REDS target
    detectors (R, E, D, and S). The dotted vertical line indicates the value
    adopted in the full configuration.}
    \label{fig:ablation_sweeps}
\end{figure*}

\begin{table*}[t]
\centering
\caption{Module ablation on REDS. Each cell reports ASR$\uparrow$ (\%) / BRISQUE$\downarrow$ for the full method and four ablated variants (defined in the text). Bold: best per column. Full (Ours) is the same configuration reported in Table~\ref{tab:blackbox_attack}.}
\label{tab:module_ablation}
\begin{tabular}{lcccccccccc}
\toprule
\multirow{2}{*}{\textbf{Config}} & \multicolumn{2}{c}{\textbf{R}} & \multicolumn{2}{c}{\textbf{E}} & \multicolumn{2}{c}{\textbf{D}} & \multicolumn{2}{c}{\textbf{S}} & \multicolumn{2}{c}{\textbf{\textit{Avg}}} \\
\cmidrule(lr){2-3}\cmidrule(lr){4-5}\cmidrule(lr){6-7}\cmidrule(lr){8-9}\cmidrule(lr){10-11}
 & ASR & BRISQUE & ASR & BRISQUE & ASR & BRISQUE & ASR & BRISQUE & ASR & BRISQUE \\
\midrule
\rowcolor{gray!12}
\textbf{Full (Ours)} & \textbf{100.00} & 20.43 & \textbf{100.00} & 21.28 & \textbf{100.00} & 19.18 & \textbf{100.00} & 20.05 & \textbf{100.00} & 20.24 \\
\midrule
w/o SGP & 97.50 & 19.18 & 96.67 & 19.95 & 97.48 & 18.38 & 82.50 & 18.18 & 93.54 & 18.92 \\
w/o SGDS & 98.33 & 53.27 & 73.33 & 52.22 & 88.03 & 52.34 & 71.67 & 52.16 & 82.84 & 52.50 \\
w/o Momentum & 95.83 & \textbf{17.62} & 91.54 & \textbf{17.82} & 89.92 & \textbf{17.11} & 75.83 & \textbf{17.82} & 88.28 & \textbf{17.59} \\
w/o Freq. & 99.17 & 28.89 & 100.00 & 25.54 & 99.16 & 24.59 & 99.17 & 22.61 & 99.38 & 25.41 \\
\bottomrule
\end{tabular}
\end{table*}

\begin{figure*}
\centering
\includegraphics[width=0.98\linewidth]{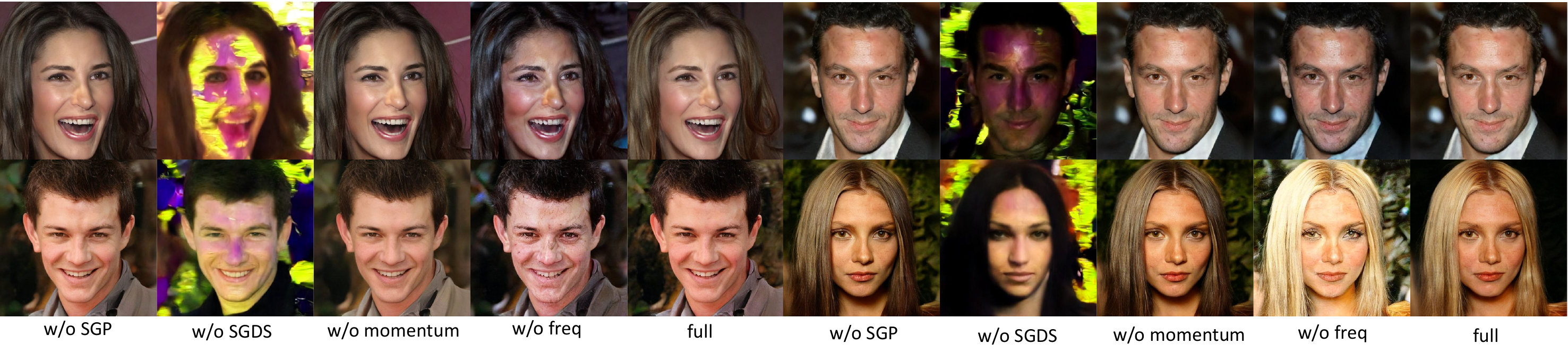}
\caption{Qualitative module ablation. Each row shows the same (mask, text) condition attacked with one component removed, alongside the full method.}
\label{fig:module_ablation}
\end{figure*}

\subsubsection{Hyperparameter Sensitivity}

Fig.~\ref{fig:ablation_sweeps} sweeps each of the five key hyperparameters independently, holding the other four at the full configuration ($\lambda_0=0.35$, $\beta_{\mathrm{guide}}=0.5$, $N=10$, $\tau_q=1$, $\beta_{\mathrm{momentum}}=0.8$). Across all five, ASR rises (near-)monotonically and then saturates while BRISQUE degrades gradually in the same direction, confirming that each parameter trades attack strength against perceptual quality as expected: $\lambda_0$ scales the injection magnitude (Eq.~\eqref{eq:inject}) and has the most direct effect on both curves; $\beta_{\mathrm{guide}}$ biases probing toward the prior (Eq.~\eqref{eq:guidedist}), raising ASR at little quality cost until orthogonal exploration is lost near $1$; larger $N$ reduces estimation variance (Eq.~\eqref{eq:dt}) with diminishing returns; smaller $\tau_q$ accumulates stronger momentum at a higher query budget; and $\beta_{\mathrm{momentum}}$ stabilizes the direction up to a point beyond which it over-smooths. In every case the adopted operating point (dotted line) sits just past the knee where ASR has saturated but before BRISQUE degrades sharply, which is why it is chosen as the full configuration.

\subsubsection{Module Ablation}

To isolate the contribution of each of the three modules described in Section~\ref{sec:method}, Table~\ref{tab:module_ablation} compares the full method against four ablated variants, all evaluated on the same four REDS targets under the same full-configuration hyperparameters: \textbf{w/o SGP} disables the surrogate-guided prior used to bias the anisotropic probing distribution described in Section~\ref{sec:method}, so the guided search falls back to using only orthogonal, isotropic exploration; \textbf{w/o SGDS} disables the zero-order directional search entirely, so the trajectory is driven only by the schedule-aware injection of whatever direction the surrogate prior alone provides; \textbf{w/o Momentum} removes momentum accumulation across guided steps by setting $\beta_{\mathrm{momentum}}=0$ in Eq.~\eqref{eq:momentum}, so each step's direction estimate is used without carrying information from previous steps; and \textbf{w/o Freq.} removes the frequency-domain artifact-suppression term of Eq.~\eqref{eq:freq} from the schedule-aware injection, leaving the raw, unfiltered update.

Table~\ref{tab:module_ablation} quantifies each component and Fig.~\ref{fig:module_ablation} shows the qualitative effect. Removing \textbf{SGDS} causes by far the largest degradation, in ASR (average $82.84\%$) and especially quality (BRISQUE $20.24\!\rightarrow\!52.50$), appearing as broad low-frequency color shifts: the zero-order search is the primary mechanism for a strong attack without a large quality-degrading injection. Removing \textbf{SGP} gives the second-largest ASR drop (average $93.54\%$, Swin-T $82.50\%$) while slightly improving BRISQUE ($18.92$), so the prior mainly contributes cross-target reliability rather than quality. Removing \textbf{Momentum} costs ASR (average $88.28\%$) at the best BRISQUE of any row ($17.59$), i.e., it trades a little quality for a more reliable attack. Removing \textbf{Freq.} barely affects ASR ($99.38\%$) but worsens BRISQUE ($25.41$), confirming that it suppresses high-frequency artifacts rather than driving attack strength. The four components are thus complementary: SGDS and SGP drive effectiveness, Momentum and Freq.\ govern the quality/reliability trade-off, and the full combination is the only configuration attaining $100\%$ ASR on all four targets while keeping BRISQUE competitive (the two variants with marginally lower BRISQUE reach it only by sacrificing ASR).

\section{Conclusion}

This paper proposed TIGA, a training-free, trajectory-injected framework that evades black-box AIGC detectors by steering the DDIM denoising trajectory through a surrogate-guided prior, an anisotropic black-box directional search, and a schedule-aware, frequency-reshaped injection. On general-purpose AIGC detectors, TIGA attains $100\%$ black-box ASR with the lowest BRISQUE among all compared methods, transfers markedly better to unseen detectors, and remains robust under common post-processing, with the module ablation confirming that its three components are complementary. Future work will include broadening the surrogate/detector pools and extending TIGA beyond face generation and a single diffusion backbone.

\bibliographystyle{IEEEtran}
\bibliography{IEEEabrv, ref}

\vfill

\end{document}